\documentclass[final,3p,times]{elsarticle}
\usepackage{amssymb}

\usepackage{tabularx}   % for auto-width tables

\usepackage{makecell}
\usepackage{comment}
 \usepackage{amsmath}
 \usepackage{booktabs}
\usepackage{multicol}
\usepackage{multirow}
\usepackage{placeins}
\usepackage{subcaption}
\usepackage{stfloats}
\usepackage{float}
\usepackage[colorlinks=true,
            linkcolor=blue]{hyperref}
\usepackage[font=normalsize,skip=3pt]{caption} 
\makeatletter
\def\ps@pprintTitle{%
  \let\@oddhead\@empty
  \let\@evenhead\@empty
  \let\@oddfoot\@empty
  \let\@evenfoot\@oddfoot}
\makeatother

\begin{document}

\begin{frontmatter}



\title{Automated Labeling of Intracranial Arteries with Uncertainty Quantification Using Deep Learning}

% --- Authors with numeric footnote refs
\author{Javier Bisbal\corref{cor1}\fnref{a1,a2,a3,a4,a8}}
\ead{jebisbal@uc.cl}
\author{Patrick Winter\fnref{a8,a10}}
\author{Sebastian Jofre\fnref{a5}}
\author{Aaron Ponce\fnref{a4,a6}}
\author{Sameer A. Ansari\fnref{a10,a11,a12}}
\author{Ramez Abdalla\fnref{a10,a11}}
\author{Michael Markl\fnref{a10}}
\author{Oliver Welin Odeback\fnref{a1}}
\author{Sergio Uribe\fnref{a7}}
\author{Cristian Tejos\fnref{a2,a3,a4}}
\author{Julio Sotelo\fnref{a5}}
\author{Susanne Schnell\corref{contrib}\fnref{a8,a10}}
\author{David Marlevi\corref{contrib}\fnref{a1,a9}}

\cortext[contrib]{Authors contributed equally.}
\cortext[cor1]{Corresponding author.}

% \fntext[eq]{These authors contributed equally to this work.}

% --- Affiliations as plain footnotes (one line each, no special macros)
\fntext[a1]{Dept. of Molecular Medicine and Surgery, Karolinska Institutet, Stockholm, Sweden}
\fntext[a2]{Biomedical Imaging Center, Pontificia Universidad Católica de Chile, Santiago, Chile}
\fntext[a3]{Department of Electrical Engineering, School of Engineering, Pontificia Universidad Católica de Chile, Santiago, Chile}
\fntext[a4]{Millennium Institute for Intelligent Healthcare Engineering (iHEALTH), Santiago, Chile}
\fntext[a5]{Departamento de Informática, Universidad Técnica Federico Santa María, Santiago, Chile}
\fntext[a6]{Escuela de Ingeniería Civil Informática, Universidad de Valparaíso, Valparaíso, Chile}
\fntext[a7]{Department of Medical Imaging and Radiation Sciences, Faculty of Medicine, Nursing and Health Sciences, Monash University, Melbourne, Australia}
\fntext[a8]{Department of Medical Physics, Institute of Physics, University of Greifswald, Greifswald, Germany}
\fntext[a9]{Institute for Medical Engineering and Science, Massachusetts Institute of Technology, Cambridge, MA, USA}
\fntext[a10]{Department of Radiology, Northwestern University, Chicago, IL, USA}
\fntext[a11]{Department of Neurological Surgery, Northwestern University, Chicago, IL, USA}
\fntext[a12]{Department of Neurology, Northwestern University, Chicago, IL, USA}

\begin{abstract}
Accurate anatomical labeling of intracranial arteries is essential for cerebrovascular diagnosis and hemodynamic analysis but remains time-consuming and subject to interoperator variability. We present a deep learning-based framework for automated artery labeling from 3D Time-of-Flight Magnetic Resonance Angiography (3D ToF-MRA) segmentations (n=35), incorporating uncertainty quantification to enhance interpretability and reliability. We evaluated three convolutional neural network architectures: (1) a UNet with residual encoder blocks, reflecting commonly used baselines in vascular labeling; (2) CS-Net, an attention-augmented UNet incorporating channel and spatial attention mechanisms for enhanced curvilinear structure recognition; and (3) nnUNet, a self-configuring framework that automates preprocessing, training, and architectural adaptation based on dataset characteristics. Among these, nnUNet achieved the highest labeling performance (average Dice score: 0.922; average surface distance: 0.387 mm), with improved robustness in anatomically complex vessels. To assess predictive confidence, we implemented test-time augmentation (TTA) and introduced a novel coordinate-guided strategy to reduce interpolation errors during augmented inference. The resulting uncertainty maps reliably indicated regions of anatomical ambiguity, pathological variation, or manual labeling inconsistency. We further validated clinical utility by comparing flow velocities derived from automated and manual labels in co-registered 4D Flow MRI datasets, observing close agreement with no statistically significant differences. Our framework offers a scalable, accurate, and uncertainty-aware solution for automated cerebrovascular labeling, supporting downstream hemodynamic analysis and facilitating clinical integration. 

\end{abstract}

\begin{keyword}
 Intracranial artery labeling\sep 3D ToF-MRA \sep Deep learning  \sep UNet \sep Intracranial 4D Flow MRI
\end{keyword}

\end{frontmatter}


\section{Introduction}
The intracranial arterial system plays a critical role in brain perfusion to maintain normal cognitive function. Occlusion or stenosis of these blood vessels can cause vascular alterations that contribute to the development of cerebrovascular or neurodegenerative diseases \cite{hinman2017principles,morgan20214d}. 

Three-dimensional Time-of-Flight Magnetic Resonance Angiography (3D ToF-MRA) is the clinical gold standard for non-invasive imaging of the intracranial vasculature. Recently, 4D Flow MRI has emerged as a new modality, adding valuable functional hemodynamic data including regional blood flow variations. In particular, intracranial 4D Flow MRI has shown promise in assessing a variety of vascular pathologies, including aneurysms \cite{hope2010evaluation,schnell2014three,liu2018highly}, arteriovenous malformations \cite{hope2009complete,aristova2019standardized}, and intracranial atherosclerotic disease (ICAD) \cite{vali2019semi,el2025non}.

Accurate quantification of 4D Flow MRI data is highly dependent on both segmentation and precise anatomical labeling of the intracranial arteries \cite{alpers1959anatomical,chen2019quantification}. Although several methods have been proposed for automated segmentation of the vascular tree \cite{winter2024automated,mou2021cs2,dou20173d,chen2018voxresnet}, our study focuses on automated labeling of the major vascular structures. This process remains one of the most labor intensive and clinically critical tasks in cerebrovascular imaging and is often significantly affected by interoperator variability \cite{chen2019quantification}.

 To achieve automated vessel labeling, traditional graph-based approaches model the vascular tree using relational graphs derived from the centerlines \cite{sobisch2022automated,chen2020automated,zhu2022deep}. However, their performance is often compromised in cases of severe stenosis or disconnected vasculature. Moreover, these methods typically overlook contextual information embedded in the 3D image space.

Deep learning (DL) models offer an alternative by learning vessel features directly from image data. These models leverage structural and spatial context to produce anatomically consistent labeling results. Most of the current literature focuses on variants of the UNet architecture \cite{hilbert2022anatomical,lv2023deep,dumais2022eicab,chen2024automated,colombo2025accuracy}, which have shown strong baseline performance. However, the field lacks standardized procedures for selecting and designing network architectures, which directly affect performance and generalization \cite{isensee2021nnu}. Additionally, advanced architectural components, such as spatial attention mechanisms, which have shown benefits in segmenting curvilinear structures \cite{mou2021cs2}, have not been explored in intracranial artery labeling. Furthermore, current approaches do not incorporate uncertainty quantification, limiting explainability and adoption in clinical settings.

To address these challenges, we performed an evaluation of UNet-based architectures to label intracranial arteries using 3D ToF-MRA, focusing on three key objectives. First, to address the lack of standardization, we leverage the self-configuring nnUNet framework, which automatically tailors the network architecture and training pipeline to the dataset \cite{isensee2024nnu}. Second, we investigate architectural refinements by adapting and assessing spatial attention mechanisms, previously applied to the segmentation of curvilinear structures, to intracranial artery labeling and evaluate their potential to improve anatomical labeling \cite{mou2021cs2}. Third, to enable uncertainty-aware predictions, we incorporate test-time augmentation (TTA) \cite{wang2019aleatoric} and introduce a novel coordinate-guided strategy to reduce interpolation errors during inference, thus improving the reliability of uncertainty estimates.

% In this study, we conduct an evaluation of UNet-based architectures for labeling intracranial arteries using 3D ToF-MRA, with a focus on three key objectives. First, to address the lack of standardized procedures for network selection and training configuration, we evaluated the self-configuring "no-new" UNet (nnUNet) framework \cite{isensee2024nnu}. This algorithm automatically tailors preprocessing steps, training settings, and architectural design to the specific characteristics of the input data, improving model generalization and performance.

% Second, we investigate the role of architectural refinements in improving labeling accuracy. Specifically, we examine the impact of spatial attention mechanisms previously applied to segmentation tasks involving curvilinear structures \cite{mou2021cs2}. Here, we adapt and assess these attention modules for their potential to improve anatomical consistency in labeling.

% Third, to enable uncertainty-aware predictions, we incorporate test time augmentation (TTA) to estimate uncertainty in the labeling output \cite{wang2019aleatoric}. We also introduce a novel coordinate-guided TTA strategy, designed to reduce interpolation errors during augmented inference. 

Together, these contributions aim to build a more robust and scalable framework for automated labeling of intracranial arteries, which is a critical step for subsequent advanced analyzes, including hemodynamic assessments using 4D Flow MRI.

% facilitating hemodynamic assessments using 4D Flow MRI.

\section{Methods}\label{Methods}

\subsection{Study cohort}

% \julio{This retrospective study included 25 patients (11 females) }
%The study retrospectively included XX patients 

We retrospectively selected 25 patients (11 females) from an IRB-approved ICAD study at Northwestern Memorial Hospital. %study, with subjects originally recruited with informed written consent (n=25; n=11 women). n=14 
Fourteen cases exhibited severe stenosis, defined as constriction $>$ 70\%, while the remaining cases showed moderate stenosis, with constriction ranging between 50\% and 70\%. The affected vessels included the middle cerebral arteries (MCAs), internal carotid arteries (ICAs), and the basilar artery (BA). %from $\geq$ 50\% to $\leq$ 70\%. 
Two interventional neuroradiologists (RA, SA) reviewed the clinical electronic medical records from MRI/MRA and MR vessel wall imaging to confirm ICAD-related stenoses.

%To complement the above, 
We also included data from ten healthy volunteers (six females). %(n=10, n=6 women). 
Informed written consent was obtained from all participants in this study. A summary of demographic and physiological data for patients and volunteers is shown in \autoref{TAB:1}.

\begin{table}[!h]
    \centering
    \caption{Median age, BMI, and average heart rate (maximum and minimum values) for the Control and ICAD groups.}\label{TAB:1}
    \begin{tabular}{lcc}
        \hline
        Parameter & Controls (n=10)& ICAD (n=25)\\
        \hline
        Age (years)          & 27  & 64  \\ 
                             & (19--35)  & (34--85) \\  
        BMI (kg/m$^2$)       & 30.30  & 27.99  \\ 
                             & (19.37--38.73)  & (21.91--41.29) \\  
        Heart Rate (bpm)     & 90.6  & 73.8  \\ 
                             & (72.6--121.2)  & (63.0--115.2) \\  
        \hline
    \end{tabular}
\end{table}

\subsection{MRI Acquisitions}  

Patients and volunteers were scanned using a clinical MRI protocol designed for intracranial vascular assessment. This protocol included a gradient-echo 3D ToF-MRA sequence to visualize vascular anatomy, and an ECG-triggered intracranial 4D Flow MRI sequence to capture blood flow dynamics. The 4D Flow MRI sequence used a dual-velocity encoding acquisition (dual-VENC) and was accelerated using PEAK GRAPPA with an acceleration factor of R = 5, as described in \cite{schnell2017accelerated}. All scans were performed at 3T (Siemens MAGNETOM Skyra, Erlangen, Germany; and Siemens MAGNETOM Prisma Fit) using a 20-channel head/neck coil (Siemens, Erlangen, Germany). Detailed scan parameters for 3D ToF-MRA and 4D Flow MRI are provided in \autoref{TAB:2}.

\begin{table}[h]
    \centering
    % \footnotesize
    \caption{Scan parameters for 3D ToF-MRA and dual-VENC 4D Flow MRI for Control and ICAD groups.}
    \label{TAB:2}
    \begin{tabular}{lcc}
        % \hline
        \multicolumn{3}{l}{\textbf{3D ToF-MRA Parameters}} \\
        \hline
        Parameters & Controls & ICAD\\
        TR [ms] & 22 & 21 \\
        TE [ms] & 3.42 & 3.4 \\
        Voxel size [mm] & 0.52 & 0.52 \\
        Slice thickness [mm] & 0.50 & 0.50 \\
        Flip angle [°] & 17 & 17 \\
        Scan time [min] & 4--5 & 4--5 \\
        Siemens MR system & Prisma Fit & Skyra \\
        \hline
        \noalign{\vskip 2mm} % Separation between blocks
        \multicolumn{3}{l}{\textbf{Dual-VENC 4D Flow MRI Parameters}} \\
        \hline
        Parameters & Controls & ICAD \\
        TR [ms] & 5.9 & 6.1--6.2 \\
        TE [ms] & 3.25 & 3.4 \\
        Temporal resolution [ms] & 82.6 & 42.7 - 86.8 \\
        Voxel size [mm] & 0.982 & 0.978--1.146 \\
        Slice thickness [mm] & 1.0 & 1.0--1.2 \\
        Number of slices & 44 & 40--60 \\
        Number of cardiac phases & 5--9 & 5--18 \\
        Flip angle [°] & 15 & 15 \\
        Low venc/high venc [m/s] & \makecell{0.5--0.6 /\\1.0--1.2} & \makecell{0.5--0.6 /\\1.0--1.2} \\
        Siemens MR system & Prisma Fit & Skyra \\
        \hline
    \end{tabular}
\end{table}

\subsection{Automated labeling}

\subsubsection{Data preparation}

\begin{figure}
	\centering
	\includegraphics[width=.5\linewidth]{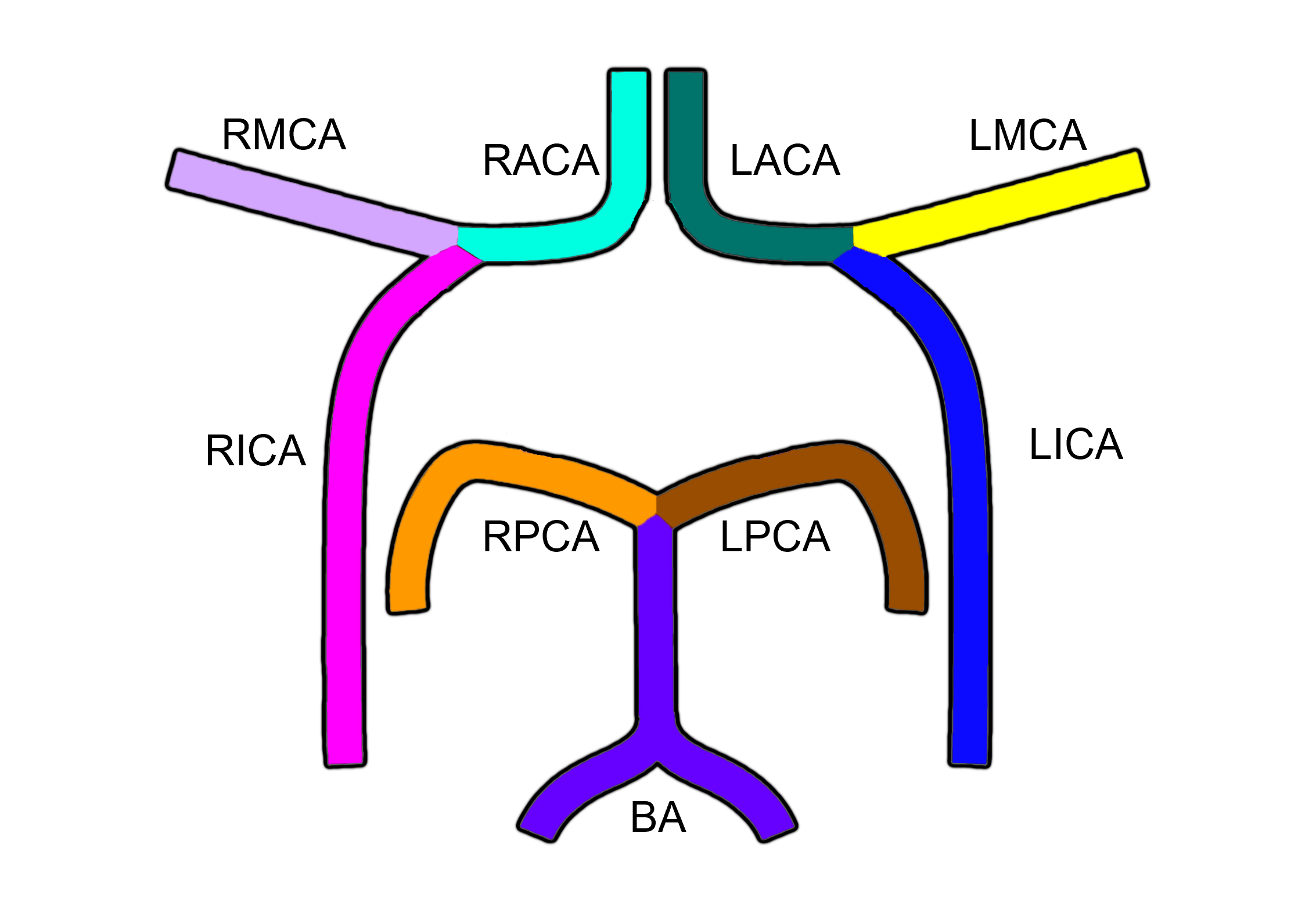}
	\caption{Schematic representation of the nine intracranial artery segments included in this study: basilar artery (BA); right and left internal carotid arteries (RICA, LICA); right and left middle cerebral arteries (RMCA, LMCA); right and left anterior cerebral arteries (RACA, LACA); and right and left posterior cerebral arteries (RPCA, LPCA).}
	\label{FIG:1}
\end{figure}

We generated binary masks of the cerebral vasculature using semi-automatic thresholding applied to the 3D ToF-MRA images. As in previous work \cite{vali2019semi}, an in-house algorithm was employed to automatically extract the centerlines of the cerebral vasculature. The identified centerlines were then manually labeled according to the anatomical section of the vessels. We focused our study on annotating nine major vessels of the Circle of Willis: the Basilar Artery (BA), Right and Left Internal Carotid Arteries (RICA and LICA), right and left middle cerebral arteries (RMCA and LMCA), Right and Left Anterior Cerebral Arteries (RACA and LACA), and Right and Left Posterior Cerebral Arteries (RPCA and LPCA). These vessels are illustrated in \autoref{FIG:1}.

To generate voxel-wise ground truth labels, we used a 7×7×7 voxel neighborhood around each centerline position. Each voxel within this neighborhood was assigned the label of its closest centerline point. This procedure produced multi-class masks for each binary ToF-MRA segmentations. Voxels without a centerline point within their 7×7×7 neighborhood were classified as "non-annotated".

% To generate ground truth in an voxel-wise level, we used a neighborhood of 7x7x7 voxels around each centerline position. We then assigned neighbor voxels the label of the closest centerline position. With this procedure we generate multi-class masks for every ToF-MRA binary segmentation.

\subsubsection{Network architectures}

Our work evaluated three state-of-the-art UNet variants for semantic segmentation, each chosen with a distinct objective: 1) automate preprocessing, hyperparameter tuning, and training strategies; 2) integrate advanced architectural refinements; and 3) benchmark against previous work.

\begin{enumerate}[1.]

    \item \textbf{Self-configuring UNet} \cite{isensee2024nnu} (nnUNet): nnUNet is a self-configuring framework that automatically optimizes the UNet with a residual encoder architecture and a training pipeline based on the input dataset. Unlike the other variants, nnUNet does not introduce new architectural innovations, but instead focuses on automating preprocessing, hyperparameter tuning, and training strategies. This approach ensures that the model is tailored to the specific characteristics of the dataset, often achieving state-of-the-art performance without manual intervention.
    \cite{colombo2025accuracy}
    
    \item \textbf{Channel and Spatial Attention Network} \cite{mou2021cs2} (CS-Net): CS-Net enhances the standard UNet architecture by integrating channel and spatial attention mechanisms. These mechanisms enable the network to better capture fine-grained details and contextual information, making this variant particularly well suited for tasks requiring precise segmentation of curvilinear structures. Originally designed for segmentation, we adapted its implementation for our labeling task.

    \item \textbf{UNet} \cite{kerfoot2019left} (baseline): This variant replaces the standard UNet encoder with a residual encoder, incorporating skip connections at each encoder layer to improve gradient flow and mitigate the vanishing gradient problem \cite{he2016deep}. As it represents the architecture used in recent published work on automated intracranial labeling \cite{lv2023deep,chen2024automated}, this UNet variant serves as our baseline model.
    
\end{enumerate}

\subsubsection{Preprocessing}

    Two preprocessing pipelines were implemented, ensuring that the input images aligned with the input formats required for the UNet, CS-Net, and nnUNet setups, respectively.

    \begin{enumerate}[1.]
        \item \textbf{Scaling with zero-padding and cropping (UNet and CS-Net)}:
        Convolutional networks typically require input images of a fixed size. To handle images of varying dimensions, we first extracted a bounding box that encapsulated each segmentation, adding an empirically chosen 15\% zero-padding margin along each dimension. Each image was also scaled to match the largest dimension of the target dimension, while preserving the original aspect ratio. This ensured that the anatomical structures were not distorted. Subsequently, we applied cropping to adjust the image to the exact target shape. For this application, we used a target dimension of 128\texttimes256\texttimes256 pixels.
        
        \item \textbf{Patch inference (nnUNet)}: As part of the nnUNet preprocessing pipeline, the proposed patch-based approach was utilized directly, allowing for input images of different sizes. Patches of size equal to the median shape of the dataset (80\texttimes224\texttimes160 pixels) were extracted using a sliding window strategy \cite{isensee2021nnu}. This allowed the model to predict on images of varying dimensions without resizing.
    \end{enumerate}   

\subsubsection{Loss function}

For $C$ classes and $N$ voxels, let $p_{i,c}$ be the predicted probability and $g_{i,c}\in\{0,1\}$ the one-hot ground truth. The Dice coefficient per class is
\begin{equation}
\text{Dice}(c)=\frac{2\sum_{i}^N p_{i,c} g_{i,c}}{\sum_{i}^N p_{i,c}+\sum_{i}^N g_{i,c}},
\label{eq:dice}
\end{equation}
and the Dice loss is the complement averaged over classes:
\begin{equation}
L_{\text{Dice}}=1-\frac{1}{C}\sum_{c=1}^{C}\text{Dice} (c).
\end{equation}
The cross-entropy loss is the mean over voxels and classes:
\begin{equation}
L_{\text{CE}}=-\frac{1}{N}\sum_{i=1}^{N}\sum_{c=1}^{C} g_{i,c}\log(p_{i,c}).
\end{equation}

We utilized a hybrid loss that include cross-entropy ($L_{\text{CE}}$) and Dice ($L_{Dice}$) losses with dynamic weights \cite{lv2023deep}, defined as  

\begin{equation}
\small
L=\left\{\begin{array}{cc}
L_{\text{CE}} & \text { epoch } \leq \beta \\
\alpha L_{\text{CE}}+(1-\alpha) L_{\text {Dice }} & \beta<\text { epoch } \leq \gamma \\
0.9 L_{\text {Dice }}  + 0.1L_{\text{CE}}, & \gamma<\text { epoch } \leq \text { total }
\end{array}\right.
\end{equation}

where
\begin{equation}
\alpha=0.8\left(1-\frac{\text{epoch}-\beta}{\gamma-\beta}\right)+0.1.
\end{equation}

This loss and the accompanying weights ensure that, during the initial training epochs (\( \leq\beta\)), the networks focus on minimizing the cross-entropy loss, with more stable convergence. Between epochs \(\beta\) and \(\gamma\), the contribution of the two losses is balanced by the weight factor \(\alpha\), which gradually decreases as training progresses, shifting the emphasis from cross-entropy to Dice loss. Finally, after epoch \(\gamma\), the loss stabilizes with a fixed weighting of 0.9 for Dice and 0.1 for cross-entropy, so the networks prioritize Dice loss, which, although less stable, can lead to more accurate predictions \cite{lv2023deep}.

% \begin{table*}[width=.99\textwidth,cols=10,pos=h]
% \centering
% \footnotesize
% \caption{Architecture details of UNet, CS-Net, and nnUNet. Stride and kernel size have the same value for all dimensions.}\label{TAB:4}
% \begin{tabular*}{\tblwidth}{@{} LLLLLLLLLL @{} }
% \toprule
% \textbf{Network} & \textbf{Input size (pixels)} & \textbf{Layers} & \makecell{\textbf{Channels} \\ \textbf{size}} & \textbf{Stride} & \makecell{\textbf{Kernel} \\ \textbf{size}} & \textbf{Activation} & \makecell{\textbf{Sliding} \\ \textbf{window}} & \makecell{\textbf{Residual} \\ \textbf{encoder}} & \makecell{\textbf{Spatial} \\ \textbf{attention}}  \\ 
% \midrule
% UNet    & 128x256x256 & 6 & (16, 32, 64, 128, 256, 512) & 2 & 3 & PReLU    & No     & Yes & No \\
% CS-Net  & 128x256x256 & 5 & (16, 32, 64, 128, 256)      & 2 & 3 & ReLU     & No     & Yes & Yes  \\
% nnUNet  & 80x224x160  & 6 & (32, 64, 128, 256, 320, 320) & 2 & 3 & LeakyReLU & Yes    & Yes  & No \\
% \bottomrule
% \end{tabular*}
% \end{table*}

\begin{table}[h]
\centering
\footnotesize
\caption{Architecture details of UNet, CS-Net, and nnUNet. Stride and kernel size have the same value for all dimensions.}\label{TAB:4}
\resizebox{\linewidth}{!}{\begin{tabularx}{1.01\linewidth}{@{} l c c c c c c c c c @{}}
\toprule
\textbf{Network} &
\makecell{\textbf{Input size}\\\textbf{(pixels)}} &
\textbf{Layers} &
\makecell{\textbf{Channels}\\\textbf{size}} &
\textbf{Stride} &
\makecell{\textbf{Kernel}\\\textbf{size}} &
\textbf{Activation} &
\makecell{\textbf{Sliding}\\\textbf{window}} &
\makecell{\textbf{Residual}\\\textbf{encoder}} &
\makecell{\textbf{Spatial}\\\textbf{attention}} \\
\midrule
UNet   & $128\times256\times256$ & 6 & (16, 32, 64, 128, 256, 512)  & 2 & 3 & PReLU     & No  & Yes & No \\
CS-Net & $128\times256\times256$ & 5 & (16, 32, 64, 128, 256)       & 2 & 3 & ReLU      & No  & Yes & Yes \\
nnUNet & $80\times224\times160$  & 6 & (32, 64, 128, 256, 320, 320) & 2 & 3 & LeakyReLU & Yes & Yes & No \\
\bottomrule
\end{tabularx}}
\end{table}

\begin{figure}
	\centering
	\includegraphics[width=.93\textwidth]{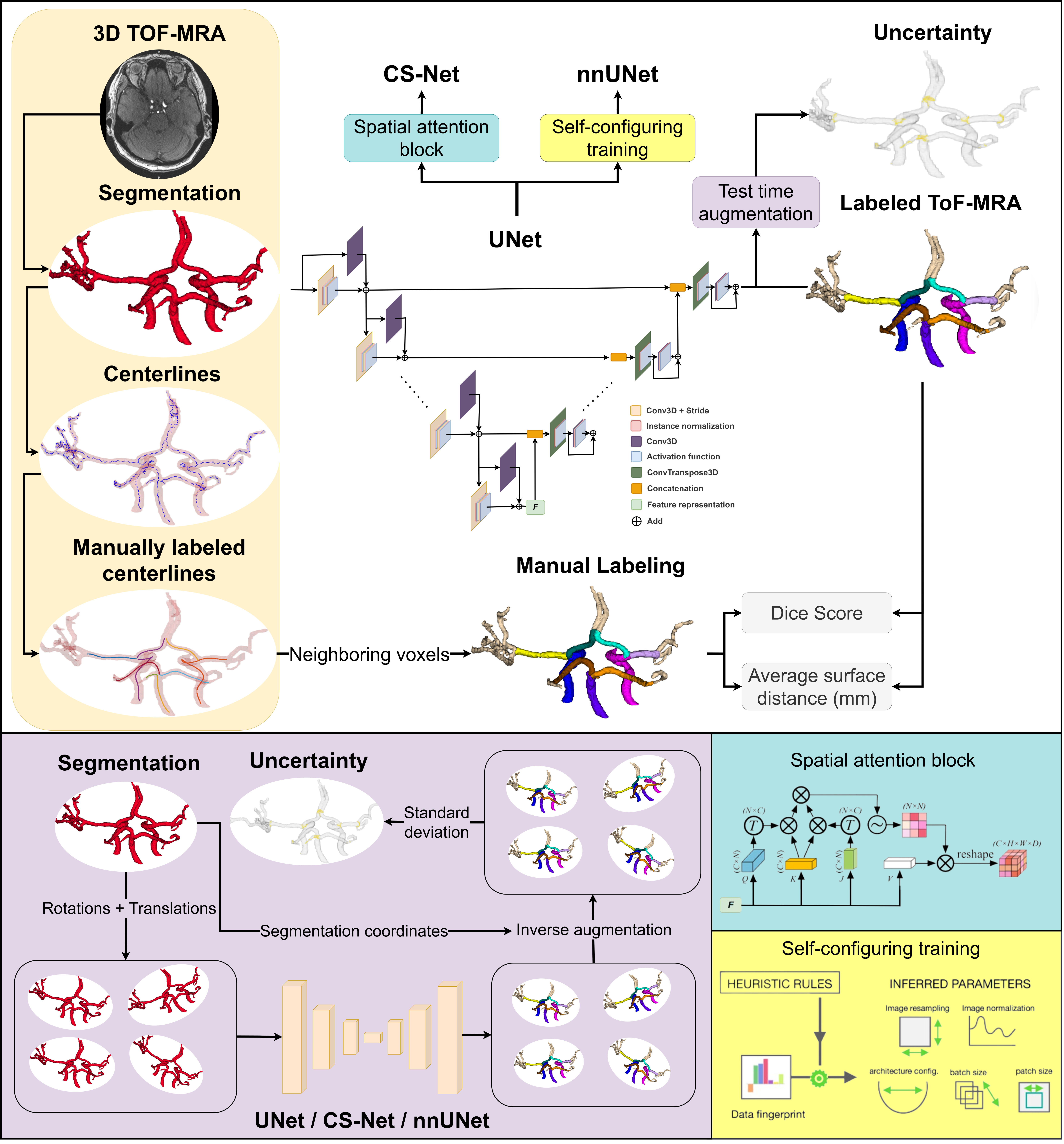}
	\caption{\textbf{Overview of the automated labeling framework}. 3D ToF-MRA images were used to segment intracranial arteries, followed by centerline extraction and manual labeling of nine arterial segments. Neighboring voxels of centerlines were labeled for voxel-wise classification. Three UNet variants (UNet, CS-Net, and nnUNet) were trained to perform voxel-wise classification directly from the segmentations. Test-time augmentation (TTA) was applied to estimate uncertainty. Labeling performance was evaluated using Dice Score and Average Surface Distance (ASD).} 
	\label{FIG:2}
\end{figure}

\subsubsection{Uncertainty quantification with modified test-time augmentation}\label{UncertatintySection}

We estimated the uncertainty using test-time augmentation (TTA) \cite{wang2019aleatoric}. TTA generates multiple slightly different inputs, each representing a plausible variation of the original data. If the labeling varies significantly across different augmented versions, it suggests that the model is less confident in its prediction for that particular region, likely due to inherent data variability arising from factors such as anatomical abnormalities, imaging artifacts, or ambiguities in ground truth annotation.

In our implementation, for each evaluation during inference, we applied random rotations (within ±18° per axis) and translations (within ±5 voxels per axis), simulating variations in the positioning of the region of interest. For each acquisition, we generated seven predictions. Then, we inverted the transformations and computed the variance across predictions as a measure of uncertainty.

Although rotations and translations are inherently invertible, interpolations are not and can introduce errors, particularly at label boundaries, which may distort uncertainty estimates using TTA. To mitigate this, we developed a coordinate-guided transformation method. For each point in the original image, we defined X, Y, and Z grids, applied the TTA transformations to these grids, and then rounded the transformed coordinates to map them back to the original space. Labels were assigned based on the closest valid segmentation value, minimizing misassignments to the background. This approach significantly reduced interpolation errors, with only minor errors remaining at the label interfaces.

We performed an experiment to prove that our coordinate-guided strategy improves the estimates of TTA. More details can be found in \ref{AppendixA}.

An overview of the automated labeling framework including preprocessing, labeling, and uncertainty quantification is shown in \autoref{FIG:2}.

\subsection{Experimental setup}

To estimate the accuracy and stability of the networks, we performed a 5-fold stratified cross-validation. For each iteration of cross-validation, 28 scans were used for training (80\% of the data), and the remaining 7 scans were used for testing (20\% of the data).

\subsubsection{Network implementation}

UNet and CS-Net were implemented using the MONAI open-source framework \cite{cardoso2022monai}, while nnUNet was implemented using its public library\footnote{\url{https://pypi.org/project/nnunetv2/}}. All implementations were built on the PyTorch deep learning framework \cite{paszke2019pytorch}.  Training, testing, and uncertainty quantification scripts, as well as model weights, are publicly available\footnote{\url{https://github.com/JavierBZ/IC-UNet}}. Note that reproducing the training for nnUNet requires following the guidelines provided in \cite{isensee2021nnu}.

UNet and CS-Net  networks were trained using the Adam optimizer \cite{kingma2014adam}, while nnUNet was trained with stochastic gradient descent (SGD). A linear learning rate scheduler was used for all optimizers, with parameters set specifically for each optimizer. For Adam, the initial learning rate was set to 0.001 and the final learning rate to 0.0001, while for SGD, the initial learning rate was 0.01 and the final learning rate was 0.001. 

UNet and CS-Net were trained for 6000 epochs with $\beta=2000$ and $\gamma=3500$, while nnUNet was trained for 1000 epochs with $\beta=400$ and $\gamma=600$. The number of epochs was chosen to ensure stable convergence in the validation curves. All networks were trained on NVIDIA A100-SXM4-40GB GPU. The total training times were 23:30 (HH:MM) for UNet, 27:30 for CS-Net, and 38:05 for nnUNet.

Specific architectural details, including convolutional layers, activation functions, and additional refinements, can be found in \autoref{TAB:4}.

\subsubsection{Automated labeling evaluation}

For evaluation, we considered the models at the last epoch of each training session. This approach prevents any bias toward better performing models in the test set for each cross-validation iteration.

Two metrics were used to evaluate the performance of the automated labeling: the Dice score (Dice) (Eq. \ref{eq:dice}) and the average surface distance (ASD), defined as:

% \begin{equation}
% D S C=\frac{2|X \cap Y|}{|X|+|Y|}
% \end{equation}

% \begin{equation}
% ASD = \frac{1}{|S_X| + |S_Y|} \left( \sum_{x \in S_X} d(x, S_Y) + \sum_{y \in S_Y} d(y, S_X) \right).
% \end{equation}
{\small
\begin{align}
\text{ASD} &= \frac{1}{|S_X| + |S_Y|} \left( \sum_{x \in S_X} d(x, S_Y) + \sum_{y \in S_Y} d(y, S_X) \right). \label{eq:asd}
\end{align}
}

Here, \(X\) and \(Y\) represent the sets of manually and automatically labeled regions, respectively. The surfaces of \(X\) and \(Y\) are denoted by \(S_X\) and \(S_Y\), and \(d(a, S)\) denotes the minimum Euclidean distance between a point \(a\) and the surface \(S\). $|.|$ computes the area of each surface. 
The Dice score measures the overlap between two sets, X and Y, and ranges from 0 (no overlap) to 1 (perfect overlap). On the other hand, ASD quantifies the average distance between the surfaces of the two sets, providing a measure of the spatial agreement between them: lower ASD values indicate better alignment between the surfaces, and vice versa.

\subsubsection{4D Flow MRI velocity analysis}
% With the idea of utilizing labelled anatomical images to guide the analysis of coupled, acquired 4D Flow datasets, assessment of 4D Flow data was also performed (seeing whether hemodynamic quantification aligned between reference manual vs. network-based labels)

To investigate whether anatomical labels could guide the analysis of corresponding 4D Flow MRI datasets, we evaluated the agreement between velocity measurements derived from manual reference labels and those generated by the best-performing automated labeling network. Specifically, we assessed whether the average velocity at peak systole within anatomically labeled regions was consistent between manual and network-based labels.

To co-register ToF-MRA images with 4D Flow MRI data, we used a semi-automatic MATLAB tool \cite{vali2019semi} that performed rigid registration with built-in functions from the SPM12 toolbox (Statistical Parametric Mapping 12) \cite{ashburner2005unified}. Following co-registration, both manually and automatically labeled ToF regions were downsampled by a factor of two to match the resolution of the 4D Flow MRI data.

Vessel-specific average velocities at peak systole, obtained from manual and automated labeling, were then compared using Bland–Altman analysis and the Wilcoxon signed-rank test for statistical evaluation.

\section{Results}\label{Results}

\begin{figure*}[!h]
	\centering
	\includegraphics[width=.99\textwidth]{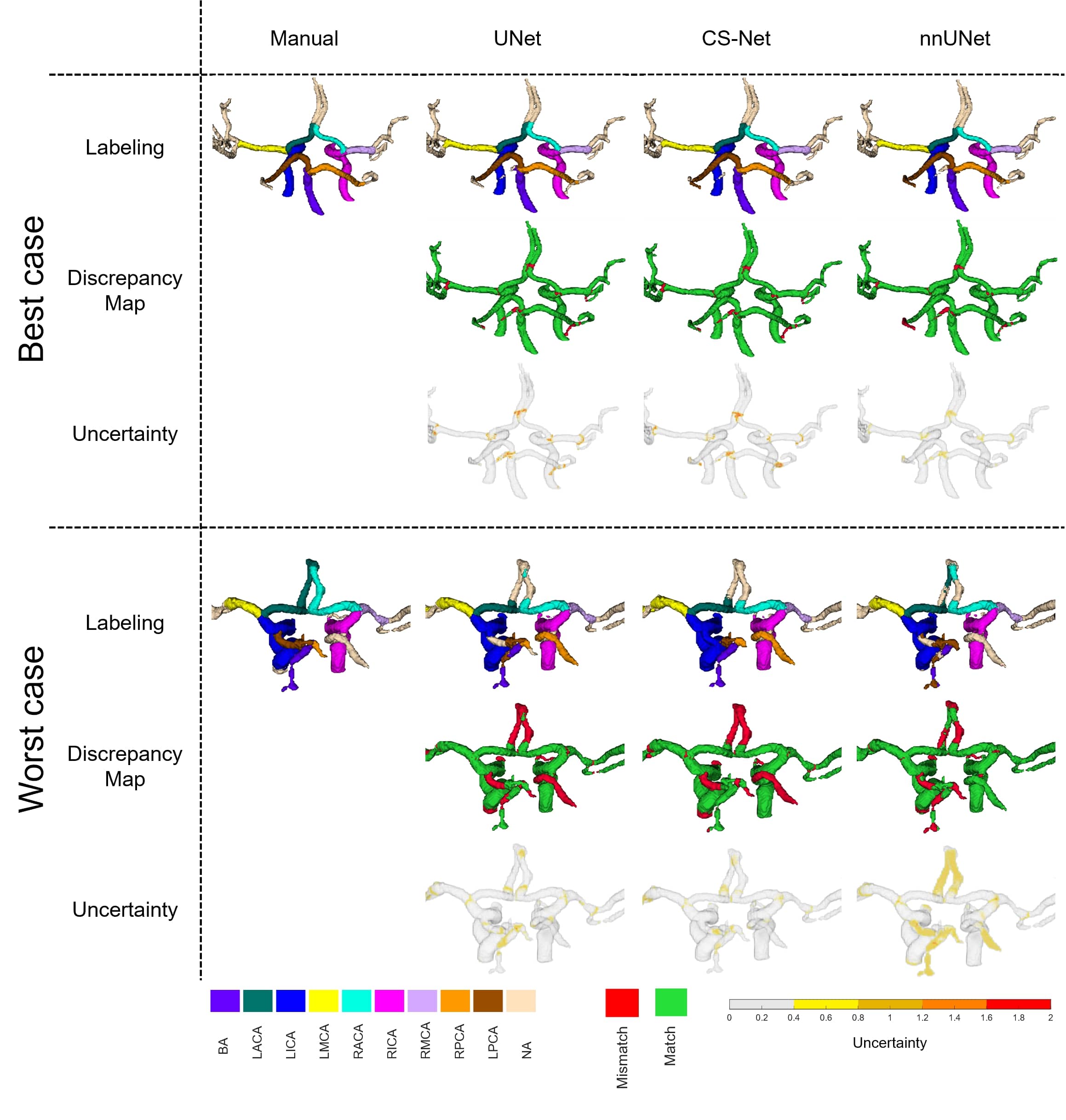}
	\caption{Automated labeling results for the best and worst-case patient data. For each case, the first row illustrates the manual labeling alongside the automated labeling predictions from UNet, CS-Net, and nnUNet. The second row highlights match and mismatch regions between the manual and automated labeling. The third row visualizes uncertainty, measured by the standard deviation of Test-Time Augmentation predictions.}
	\label{FIG:3}
\end{figure*}

\begin{figure*}[!h]
\captionsetup{type=table} % Tell LaTeX to treat this as a table
    \caption{Performance metrics of UNet, CS-Net, and nnUNet. The table presents the average Dice Score and Average Surface Distance values for different cerebrovascular segments across the three networks. Color intensity represents performance levels: lighter and darker colors indicate lower and higher performance, respectively.}
    
	\centering
	\includegraphics[width=0.8\textwidth]{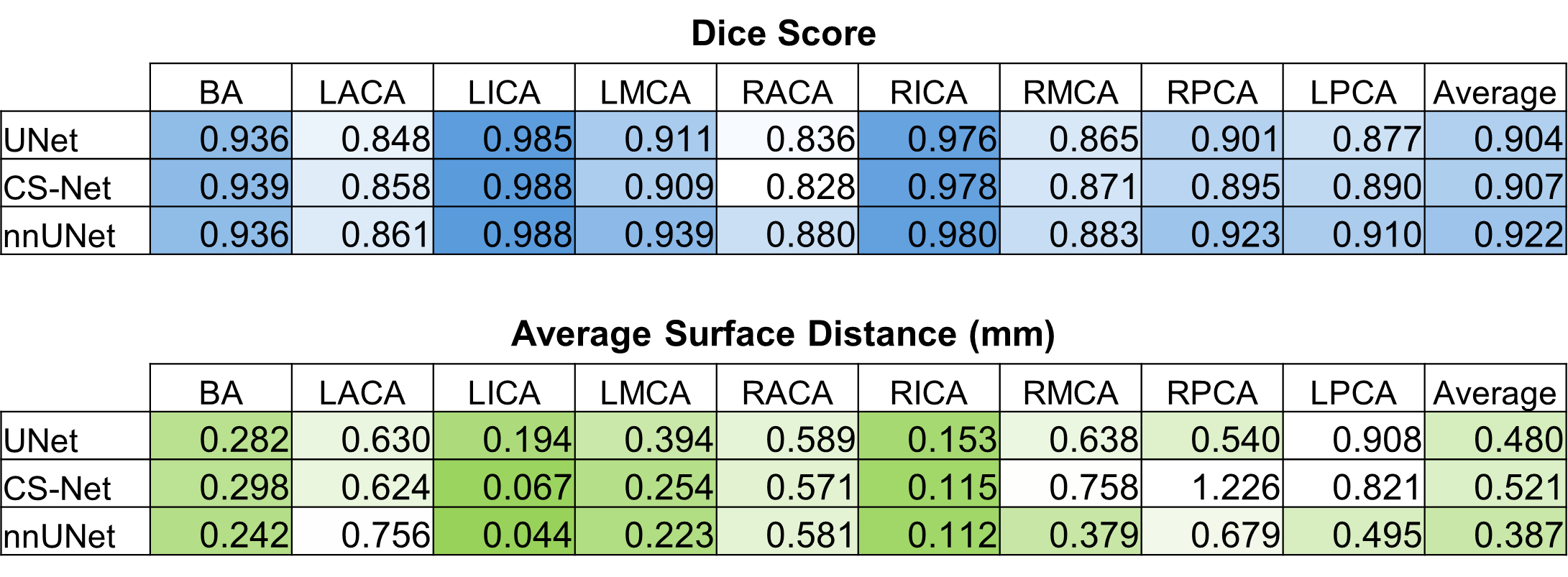}
	\label{TAB:5} % Use a label indicating it's a table
\end{figure*}
\subsection{Automated labeling}

\autoref{FIG:3} presents the automated labeling results for the best and worst cases in terms of average Dice across all networks.

In the best-case scenario all networks achieved accurate labeling of the vessels of interest with only minor errors occurring primarily at vessel bifurcations or connections with non-annotated segments. These mislabeled regions are highly correlated with areas of higher uncertainty.

In contrast, for the worst-case scenario, three regions showed some problems:
\begin{itemize}
\item \textbf{RPCA misclassification}: UNet and CS-Net incorrectly labeled the RPCA segment, probably due to its anatomical similarity and proximity to a variant vessel branching from the RICA (a common hyperplastic posterior communicating artery variant \cite{rajan2023study}). While UNet and CS-Net failed to capture this uncertainty, nnUNet correctly classified this region as non-annotated with a relatively high uncertainty, recognizing it as an anatomical variant.

\item \textbf{BA stenosis}: All networks struggled to label the BA segment in a patient with severe stenosis. UNet and CS-Net partially captured the uncertainty associated with this error, while nnUNet showed high uncertainty throughout the stenotic BA.

\item \textbf{ACA variability}: All networks misclassified upper ACA segments due to inconsistencies in manual labeling protocols, particularly to delineate terminal portions of smaller vessels. These variations were consistently reflected in the uncertainty estimates across all networks.

\end{itemize}

% \begin{figure*}
%     \captionsetup{type=table} % Tell LaTeX to treat this as a table
% 	\centering
% 	\includegraphics[width=0.9\textwidth]{figs/Table_statistics.png}
% 	\caption{Width of limits of Agreement (LoA), average absolute differences, and p-values using Wilcoxon signed-rank test for average velocity differences between manual and automated labeling.}
% 	\label{TAB:6} % Use a label indicating it's a table
% \end{figure*}

\autoref{TAB:5} shows the quantitative evaluation of the labeling performance. Consistently, strong performance is observed across all networks, with nnUNet achieving the highest overall accuracy. For Dice scores, nnUNet demonstrated superior performance (average 0.922) compared to CS-Net (0.907) and UNet (0.904), with notable improvements in the LMCA (0.939 vs. 0.911/0.909) and RACA (0.880 vs 0.836/0.828) segments. Although all networks excelled at labeling larger vessels, such as LICA and RICA (Dice $>$ 0.975), smaller vessels, including the ACA and PCA segments, proved more challenging, showing lower agreement between methods. 

\begin{figure*}[!h]
	\centering
	\includegraphics[width=\textwidth]{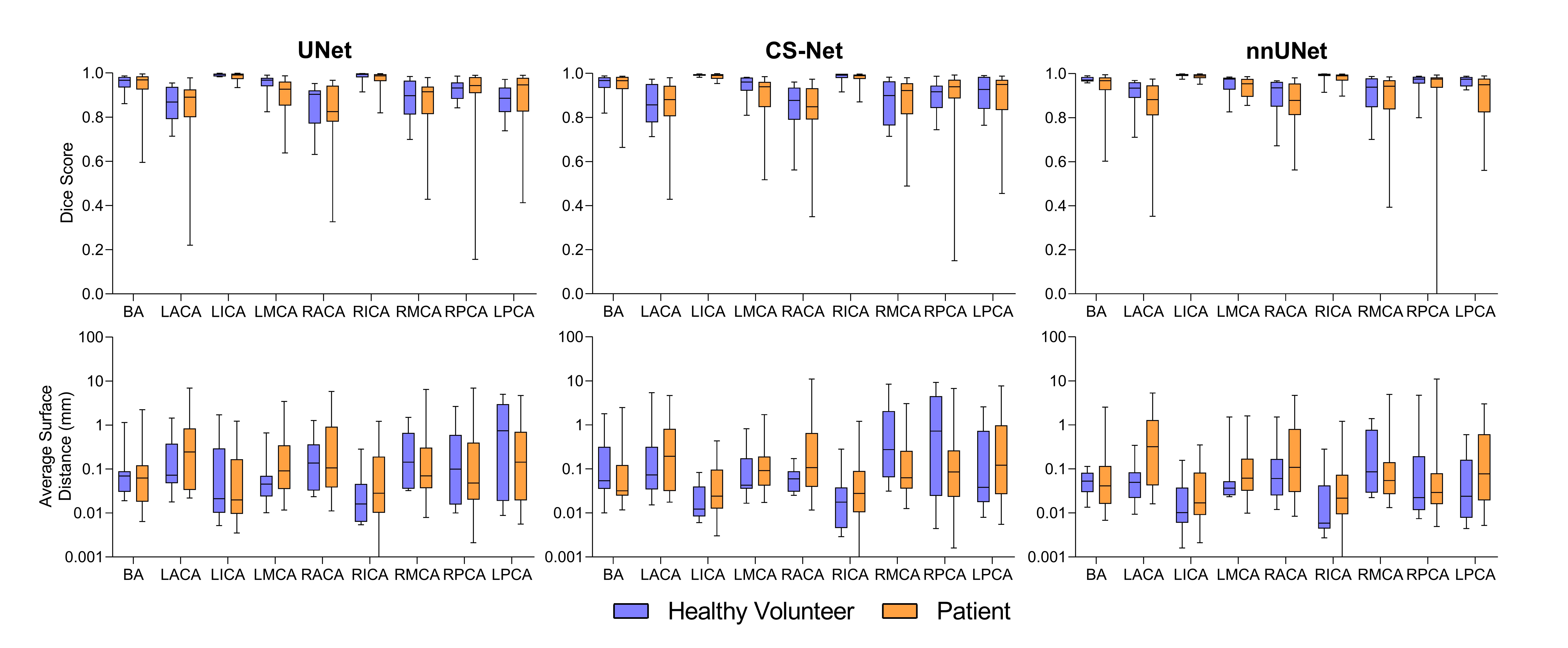}
	\caption{Boxplot analysis of Dice Score and Average Surface Distance distributions across networks comparing labeling performance between healthy volunteers (n=10) and patients (n=25). Whiskers represent the full data range (minimum to maximum values).}
	\label{FIG:4}
\end{figure*}

The lowest surface distance (which provided specific information about the boundary detection capacity) was for nnUNet (ASD = 0.387 mm) compared to UNet (0.480 mm) and CS-Net (0.521 mm). Although all networks performed well on the ICA segments ($<$0.2 mm error), nnUNet showed particular advantages in the regions RMCA (0.379 mm vs 0.638/0.758 mm) and LPCA (0.495 mm vs 0.908/0.821 mm). The BA segment showed consistent performance across networks (0.24-0.30 mm), suggesting that all architectures handled this structure effectively.

Boxplots (\autoref{FIG:4}) illustrate differences between healthy volunteers and patients, as well as between networks. For both UNet and CS-Net,the median Dice scores and ASD showed no clear differences between the healthy and patient groups, although variability tended to be higher in patients, particularly in smaller or more complex vessels such as the RPCA and LPCA. In contrast, nnUNet tended to perform worse in patients, exhibiting higher variability and worse median values for both Dice scores and ASD. Despite this, nnUNet was the most robust model overall, demonstrating the least variability across both groups.

\begin{figure*}[!h]
	\centering
	\includegraphics[width=.9\textwidth]{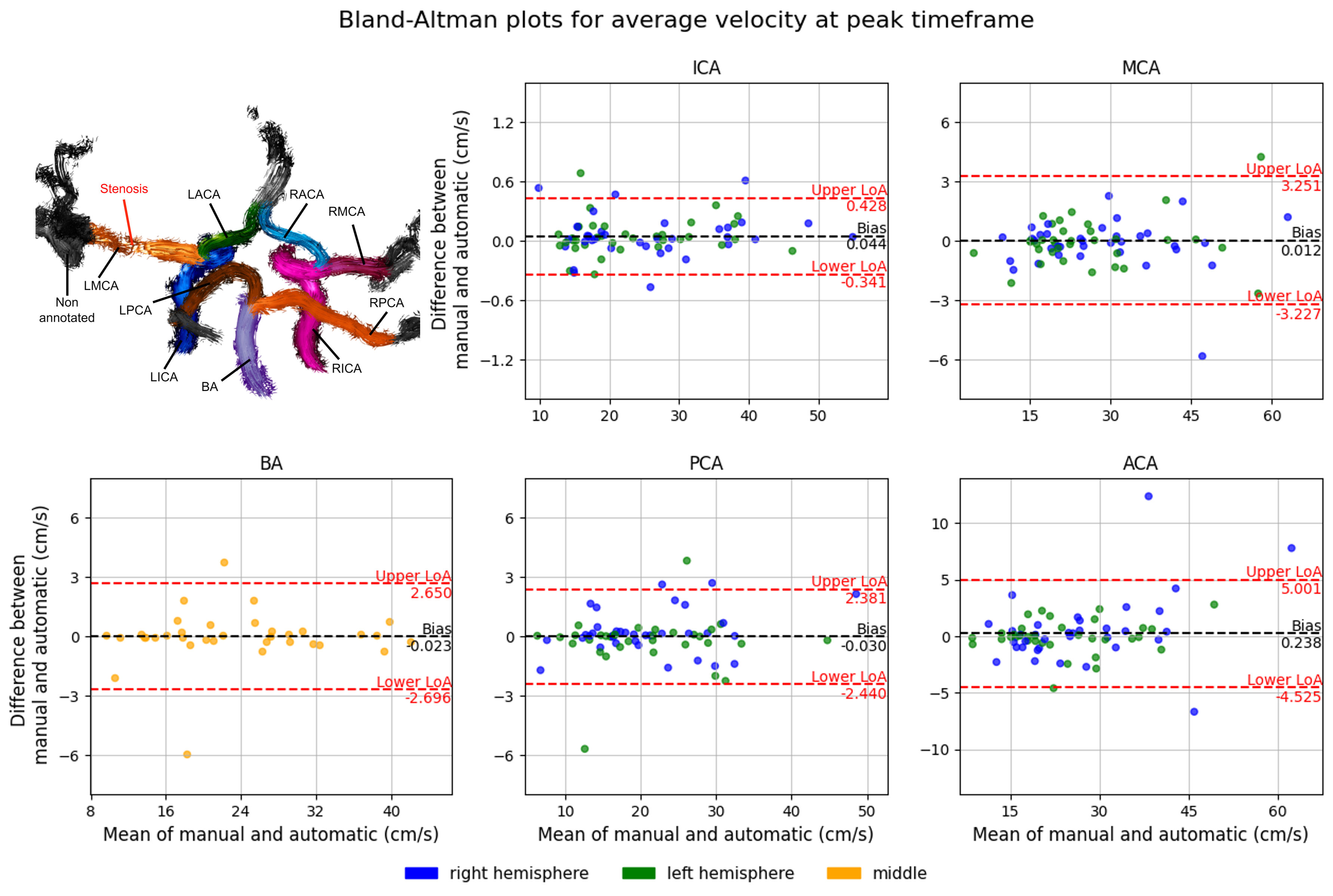}
	\caption{The upper left corner displays streamlines of different artery segments using automated labeling in a patient with severe stenosis in the left middle cerebral artery (LMCA). In the remaining figures, Bland-Altman plots compare the percentage differences in average velocity at peak systole between manual and automated labeling using nnUNet. Bias and limits of agreement are shown for each artery group: internal, middle, posterior, and anterior cerebral arteries (ICA, MCA, PCA, and ACA, respectively).}
	\label{FIG:5}
\end{figure*}

\begin{table*}[!h]
\centering
\caption{Width of limits of Agreement (LoA), average absolute differences, and p-values using Wilcoxon signed-rank test for average velocity differences between manual and automated labeling.}
\renewcommand{\arraystretch}{1.15}
\begin{tabular}{lrrrrr}
\hline
\textbf{Vessel} &
\begin{tabular}[c]{@{}r@{}}Average absolute\\ difference (cm/s)\end{tabular} &
\begin{tabular}[c]{@{}r@{}}Average absolute\\ difference (\% of mean)\end{tabular} &
\begin{tabular}[c]{@{}r@{}}Width of limits of\\ agreement (cm/s)\end{tabular} &
\begin{tabular}[c]{@{}r@{}}Width of limits of\\ agreement (\% of mean)\end{tabular} &
\textbf{p-value} \\
\hline
BA   & 0.585 & 2.490 & 4.863 & 20.704 & 0.259 \\
LACA & 1.693 & 6.306 & 12.065 & 44.931 & 0.942 \\
LICA & 0.155 & 0.605 & 0.805 & 3.142 & 0.056 \\
LMCA & 0.799 & 2.833 & 5.712 & 20.259 & 0.704 \\
RACA & 1.005 & 4.213 & 5.789 & 24.267 & 0.919 \\
RICA & 0.125 & 0.540 & 0.752 & 3.243 & 0.369 \\
RMCA & 1.000 & 3.711 & 6.209 & 23.043 & 0.968 \\
RPCA & 0.621 & 3.028 & 7.096 & 34.598 & 0.447 \\
LPCA & 0.591 & 2.828 & 4.041 & 19.342 & 0.158 \\
\hline
\end{tabular}

\label{TAB:6}
\end{table*}

% However, there are exceptions for RPCA, LACA, LPCA, and RMCA. For these vessels, volunteers showed lower median Dice for UNet and CS-Net compared to patients. This could be attributed to the smaller size and more variable anatomy of these arteries, which can make them difficult to accurately label even in healthy cases. In contrast, patients, despite having pathological changes, might exhibit more consistent labeling results for these vessels due to the networks being trained on a larger proportion of patient data.

% Regarding ASD, healthy volunteers generally show lower median values, but with higher variability in smaller vessels. For all networks the BA and RMCA in volunteers exhibit higher median ASD compared to patients. Also for UNet LPCA and RPCA, and for CS-Net RPCA, volunteers exhibit higher median ASD, and CS-Net. This suggests that, for these vessels, anatomical variability even in healthy cases make the networks struggle to delineate boundaries accurately.
\subsection{4D Flow MRI analysis}

\autoref{FIG:5} presents Bland-Altman plots for the 4D Flow MRI analysis, comparing the average velocity at the peak systole timeframe between manual and automated labeling using the best-performing network (nnUNet). ICAs demonstrated the smallest differences between manual and automatically labeled vessels, with agreement limits ranging from -0.34 cm/s to 0.43 cm/s (-2.1 \% of the mean to 2.4 \% of the mean, respectively). The BA also exhibited relatively small differences, although a few outliers widened its limits of agreement from -2.70 cm/s to 2.65 cm/s (-12.4 \% of the mean to 12.8 \% of the mean). Larger discrepancies were also observed in the smaller arterial groups including MCAs (limits of agreement: -3.23 cm/s to 3.25 cm/s, equivalent to -10.7 \% to 9.4 \% of the mean), PCAs (limits of agreement: -2.44 cm/s to 2.38 cm/s; equivalent to -15.5 \% to 14.1 \% of the mean), and ACAs (limits of agreement: -4.53 cm/s to 5.00 cm/s; equivalent to -15.9 \% to 17.6 \% of the mean).

\autoref{TAB:6} presents Bland-Altman individual limits of agreement and average differences for all intracranial segments. Overall, the average differences were less than 2 cm/s for all segments, with 7 out of 9 segments showing an average absolute difference below 1 cm/s. When expressed as a percentage of the mean velocity, the average absolute differences ranged 0.54\% (RICA) to 6.31\% (LACA). Statistical analysis revealed no significant differences between manual and automatically labeled vessels for any of the derived average velocities.

\section{Discussion}\label{Discussion}

\subsection{Highlights}
This study investigated the application of deep learning, specifically UNet-based architectures, for automated anatomical labeling of the major intracranial arteries from 3D ToF-MRA data. Networks also incorporate uncertainty quantification for subsequent 4D Flow MRI analysis. Our findings demonstrate that, while architectural refinements such as channel and spatial attention blocks in CS-Net offered improved performance compared to the baseline UNet, the self-configuring nnUNet framework yielded the most accurate and robust labeling results. Furthermore, we introduced a novel coordinate-guided test-time augmentation (TTA) strategy for estimating uncertainty, which effectively highlighted regions prone to labeling errors. Validation through 4D Flow MRI velocity analysis confirmed that the automated labels generated by the best performing network (nnUNet) provide a reliable basis for subsequent hemodynamic assessment.

During the drafting and revision of our manuscript, Colombo et al. (2025) \cite{colombo2025accuracy} published a nnUNet-based pipeline that simultaneously segments intracranial aneurysms and their parent vessels in 3D ToF-MRA. Their model achieved a median Dice score of 0.86 in 21 classes and an aneurysm detection sensitivity of 0.80. These results corroborate our choice of nnUNet as a strong baseline for cerebrovascular applications. However, their work did not compare to other UNet baselines or assess uncertainty quantification. By systematically benchmarking attention-based architectures and introducing coordinate-guided test-time augmentation, our study extends the nnUNet paradigm toward greater anatomical consistency and interpretability.

\subsection{Uncertainty quantification and clinical utility}
Beyond accuracy, the integration of uncertainty quantification is crucial for the clinical translation of deep learning models. Our implementation of TTA successfully captured regions where the model predictions were less confident. High uncertainty arises when an input image differs substantially from the types of image to which the model was exposed during training, for example, due to anatomical variations or disease-related changes. In these cases, applying geometric transformations, such as rotations or translations during TTA, results in highly variable predictions. This occurs because the models are trained to be invariant to such transformations only within the training data distribution. When the input deviates from that distribution, their predictions become less stable.

As illustrated in \autoref{FIG:3}, regions with higher uncertainty often corresponded to anatomical variations (e.g. hyperplastic posterior communicating artery variant affecting RPCA labeling), vascular alterations (e.g., BA stenosis), or inconsistencies arising from the manual labeling protocol itself (e.g., defining distal ACA segments). This provides clinicians with valuable information on the trustworthiness of automated labels in specific regions.

Moreover, our proposed coordinate-guided TTA strategy addresses interpolation errors during the inverse transformation step. By mapping transformed coordinates back to the original grid space, we minimized these artifacts, reducing label misassignments from near 5\% to $<$0.1\% compared to standard affine inversion and obtained cleaner uncertainty maps at the vessel boundaries (see \ref{AppendixA}).

\subsection{Labeling considerations}

For nnUNet, we also highlighted performance differences between healthy volunteers and patients with ICAD (\autoref{FIG:4}). For patients, nnUNet tended to obtain lower labeling performance and greater variability, particularly in smaller or more complex vessels. This is likely to be attributable to the increased anatomical complexity and variability introduced by cerebrovascular disease, such as stenoses or altered vessel morphologies, which pose a greater challenge to automated methods.

In contrast to previous approaches that operate directly on image intensity data, our pipeline performs anatomical labeling using pre-segmented vascular structures. This design choice reflects a deliberate focus on the spatial and geometric features of the vasculature, which are more relevant for accurate anatomical identification. Intensity-based features in ToF-MRA are often susceptible to artifacts, such as flow-related signal loss, inhomogeneities, or nonvascular enhancements, which can mislead the network and reduce its generalizability across diverse datasets \cite{wilcock1995problems}. By decoupling the labeling task from the raw image intensities and instead using binary segmentations as input, the network is encouraged to learn from the morphology and topology of the vessels themselves. Furthermore, excluding intensity features helps prevent the propagation of undesired biases from intensity features in the uncertainty estimates, resulting in more meaningful and spatially grounded uncertainty maps. 

\subsection{Validation for downstream 4D Flow MRI analysis}
The successful application of automated labels for 4D Flow MRI velocity analysis (\autoref{FIG:5}, \autoref{TAB:6}) demonstrates the practical utility of our approach. Bland-Altman analysis revealed good agreement between velocity measurements derived from manual labels and those from nnUNet's automated labels, with no statistically significant differences found for any vessel segment using the Wilcoxon signed-rank test. Although agreement was excellent for larger vessels such as the ICAs, slightly larger discrepancies were noted for smaller arteries (ACAs, PCAs, MCAs). This may reflect the slightly lower labeling accuracy in these segments, potential partial-volume effects, or minor residual co-registration inaccuracies. Nonetheless, the overall small average differences ($<$2 cm/s absolute difference for all segments, $<$1 cm/s for 7 out of 9 segments) suggest that the labeling of nnUNet provides a robust foundation for automated hemodynamic quantification, reducing the need for time-consuming manual delineation.

\subsection{Limitations}
Despite promising results, our study has some limitations. First, our method relies on an initial semi-automated segmentation step; any vessels missed or incorrectly segmented in this initial mask are not correctly labeled by the network. Errors in the initial segmentation will inevitably propagate to the final labeling. Second, the validation involving 4D Flow MRI is subject to potential co-registration errors between the ToF-MRA (used for labeling) and the 4D Flow MRI datasets, which could influence the velocity comparison. Third, the number of datasets (n=35) is relatively limited, covering healthy volunteers and ICAD patients from a single vendor. Although a 5-fold cross-validation provides a measure of robustness, evaluation on larger, more diverse datasets is needed to fully assess generalizability. However, large amounts of intracranial 4D Flow MRI datasets are rare, thus our study represents an already acceptable cohort.

\subsection{Future directions}

Several promising directions can be pursued to extend our work. Building on previous studies that investigated architectural modifications within the nnUNet framework \cite{mcconnell2023exploring}, future research could explore the integration of the attention mechanisms used in CS-Net, into the self-configuring nnUNet pipeline. This approach may further improve performance by combining the robust automated pipeline of nnUNet with architectural innovations. The resulting vessel labels could also serve as reliable inputs for downstream tasks, including automated centerline labeling, contributing to a fully automated pipeline for cerebrovascular hemodynamic analysis. Furthermore, applying transfer learning to adapt the labeling model directly to 4D Flow phase-contrast MRA (PC-MRA) data could eliminate the need for co-registration and potentially improve the accuracy of velocity quantification within automatically labeled vascular regions. In parallel, future validation efforts should include data from multiple scanner vendors and clinical sites to ensure broader applicability and to understand how scanner-specific differences may impact model generalization.

\section{Acknowledgments}

J.B. was funded by the National Agency for Research and Development (ANID) / Scholarship Program / DOCTORADO BECAS CHILE/2022 – 21220454, and by ANID - Millennium Science Initiative Program - ICN2021\_004. JS thanks to ANID Millennium Science Initiative Program ICN2021\_004 and Department of Medical Imaging and Radiation Sciences at Monash University. C.T. thanks Fondecyt 1231535 and ANID - Millennium Science Initiative Program - ICN2021\_004. S.S and P.W thanks to the National Institute of Health under R01HL115828 and R21NS106696.

% \section{References}\label{}


\section*{Declaration of generative AI and AI-assisted technologies in the writing process.}

During the preparation of this work the author(s) used GPT-5 in order to improve language and readability. After using this tool/service, the authors reviewed and edited the content as needed and take full responsibility for the content of the publication.

% \bibliographystyle{cas-model2-names}
\bibliographystyle{elsarticle-num} 
\bibliography{cas-refs}
% \bibliography{references}

\setcounter{figure}{0}
\appendix
\section{Appendix A: Label Inversion and Uncertainty Quantification}
\label{AppendixA}

To validate our coordinate-guided transformation method, we conducted an experiment using manual labels from one acquisition. We applied 100 random transformations (rotations: $\pm18^\circ$ per axis; translations: $\pm5$ voxels per axis), then evaluated two key aspects.

\begin{figure}
	\centering
	\includegraphics[width=.55\linewidth]{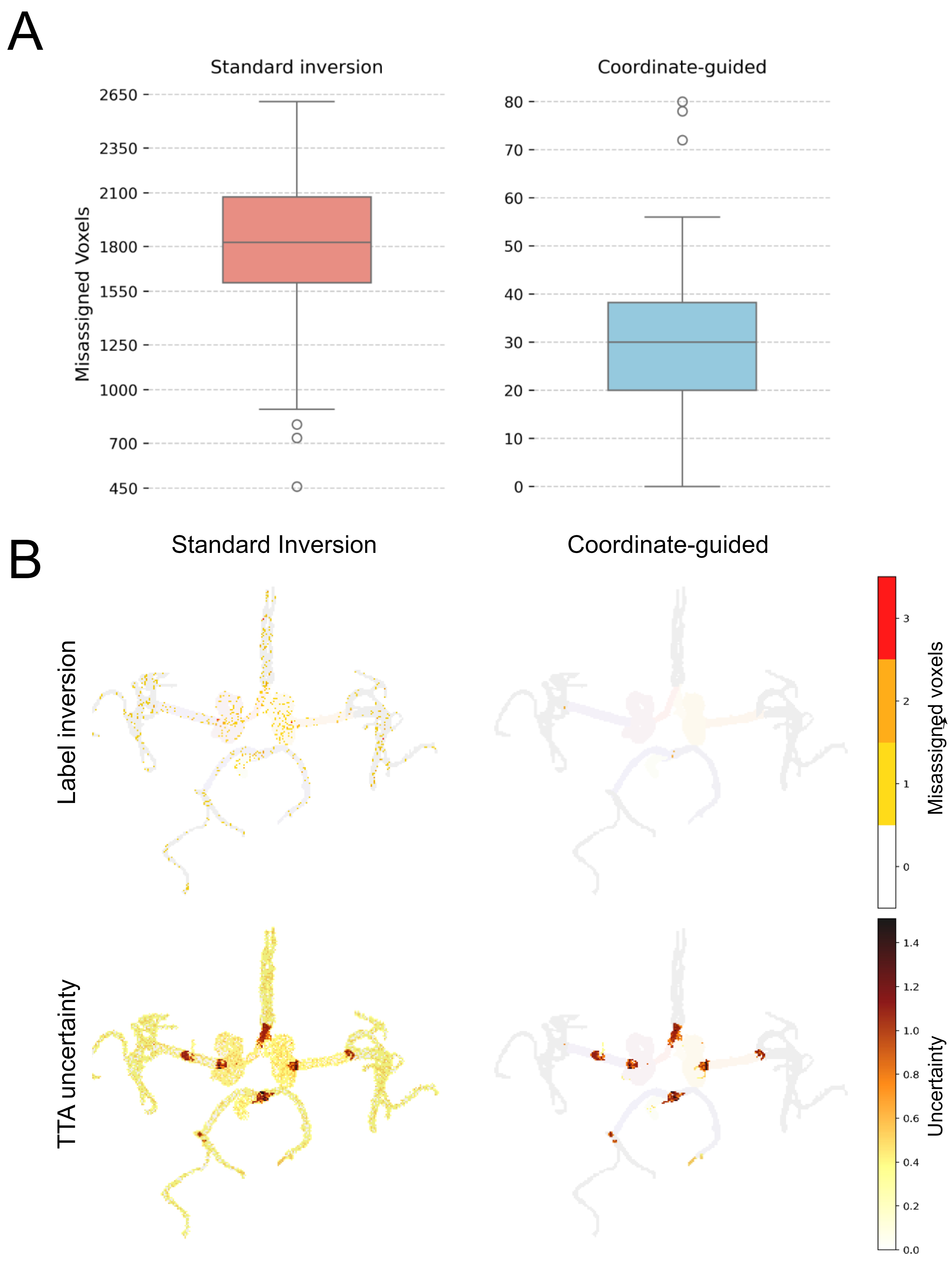}
	\caption{\textbf{A}: Distribution of label inversion errors across 100 random transformations. Left boxplot shows errors using standard affine inversion (median $5.18\%$), while right boxplot demonstrates improved performance of our coordinate-guided method (median $0.09\%$).
    \textbf{B}: Comparison of inversion methods for one random rotation and translation. First row: Label inversion results with standard inversion (left) and the coordinate-guided approach (right).The sum of misassigned voxels along the slice direction is overlaid on the maximum intensity projection of the labeled image. Second row: Uncertainty maps using standard inversion (left) and the coordinate-guided approach (right), shown as maximum intensity projections along the slice direction. }
	\label{FIG:7}
\end{figure}

First, we assessed the accuracy of label inversion by comparing standard inversion (using inverse affine matrices with nearest-neighbor interpolation) against our proposed coordinate-guided method. The proposed approach significantly reduced the inversion errors to $0.09\% \pm 0.04\%$ (mean $\pm$ standard deviation) of misassigned voxels versus $5.10\% \pm 1.17\%$ of the standard inversion. Boxplots in \autoref{FIG:7}A demonstrate that our method maintains the misassignment below $0.2\%$, while the standard inversion reaches up to $7\%$. The first row of \autoref{FIG:7}B  shows the location and amount of missassigned voxels. The second row of \autoref{FIG:7}B shows the spatial distribution of these errors: standard inversion produces misassigned voxels scattered across borders and label interfaces, while our approach yields only minimal errors at label interfaces.

Second, we examined the effects on uncertainty quantification using the nnUNet labeling network. For seven augmented versions of the acquisition, we computed prediction uncertainty maps using both inversion methods. Figure~\ref{FIG:7} (second row) shows that our method produced more reliable uncertainty estimates, particularly at segmentation boundaries, where standard interpolation introduces artifacts. Although both methods can identify high-uncertainty regions, standard inversion fails to distinguish low-uncertainty areas due to interpolation artifacts.

% \begin{figure}
% 	\centering
% 	\includegraphics[width=.99\columnwidth]{figs/double_y_violin_plot_512.png}
% 	\caption{Distribution of label inversion errors across 100 random transformations. Left violin plot shows errors using standard affine inversion (median $5.18\%$), while right plot demonstrates improved performance of our coordinate-based method (median $0.09\%$`).}
% 	\label{FIG:6}
% \end{figure}

\end{document}